\documentclass[pdflatex,sn-mathphys-num]{sn-jnl}
\usepackage{fix-cm}

\usepackage{graphicx}%
\usepackage{multirow}%
\usepackage{amsmath,amssymb,amsfonts}%
\usepackage{amsthm}%
\usepackage[title]{appendix}%
\usepackage{xcolor}%
\usepackage{textcomp}%
\usepackage{manyfoot}%
\usepackage{booktabs}%
\usepackage{algorithm}%
\usepackage{algorithmicx}%
\usepackage{algpseudocode}%
\usepackage{listings}%

\begin{document}

\title[Heterogeneity-Aware Deep Learning for Tumour Classification from Multiparametric MRI]{Heterogeneity-Aware Deep Learning for Tumour Classification from Multiparametric MRI}

\author[1]{\fnm{Yue} \sur{Xia}}

\author[1]{\fnm{Euijoon} \sur{Ahn}}

\author[2]{\fnm{Tian} \sur{Xia}}

\author[1]{\fnm{Yuan} \sur{Yuan}}

\author*[2]{\fnm{Michael} \sur{Fulham}}

\author*[1]{\fnm{Jinman} \sur{Kim}}

\affil[1]{\orgdiv{School of Computer Science}, \orgname{The University of Sydney}, \orgaddress{\city{Camperdown}, \state{NSW}, \postcode{2006}, \country{Australia}}}

\affil[2]{\orgdiv{Department of Molecular Imaging}, \orgname{Royal Prince Alfred Hospital}, \orgaddress{\city{Camperdown}, \state{NSW}, \postcode{2006}, \country{Australia}}}

\abstract{Intra-tumoural heterogeneity (ITH) reflects spatially varying biological characteristics within a tumour and is considered a major determinant of tumour behaviour, prognosis, and treatment response. Radiomics and deep learning approaches have shown promise for tumour classification using multiparametric MRI (mp-MRI). However, radiomics methods depend on handcrafted features, which may not fully capture complex tumour characteristics. Most deep learning methods instead rely on whole-tumour representations or manually defined sub-regions, limiting their ability to learn scalable and biologically meaningful representations of tumour heterogeneity. To address these limitations, we propose a Heterogeneity-Aware Deep Learning Classification (HA-DLC) framework that explicitly models imaging-derived tumour sub-regions for tumour classification tasks, including lesion-type diagnosis and molecular-status prediction. HA-DLC comprises two principal components: (1) a Heterogeneous Sub-region Generation (HSG) module that generates initial pseudo-labelled tumour sub-regions through unsupervised clustering, followed by a Cross-Patient Sub-region Alignment (CPSA) strategy that learns soft assignments from the cluster-derived regions to a shared label space; and (2) a Dual-Stream Feature Extraction (DSFE) module that integrates local heterogeneity-aware representations with global tumour features. Given the initial clustering-derived masks, the CPSA, segmentation, feature-extraction, and classification components are jointly optimized end-to-end using soft-target segmentation and classification objectives, enabling simultaneous learning of sub-regional and tumour-level information. We evaluate HA-DLC on the LLD-MMRI2023 liver lesion dataset and the RSNA-ASNR-MICCAI 2021 Radiogenomic Brain Tumour (BraTS-RC) dataset. Experimental results demonstrate that HA-DLC consistently outperforms state-of-the-art radiomics and deep learning baselines, highlighting the effectiveness of cross-patient sub-region alignment and dual-stream heterogeneity modelling for tumour classification from mp-MRI.}


\maketitle

\section{Introduction}\label{sec1}

Tumours are generally biologically heterogeneous, exhibiting spatial variation in cellular, molecular, and microenvironmental characteristics that may manifest as regional differences in their imaging phenotypes. Intra-tumoural heterogeneity (ITH) refers to these within-tumour differences, which may arise from variations in cellular composition, genomic alterations, gene expression, vasculature, necrosis, perfusion, and oxygenation, collectively reflecting the underlying tumour biology. ITH is associated with tumour behaviour, prognosis, and therapeutic response and therefore has implications for diagnosis, prognostic stratification, treatment selection, and patient management \cite{Fisher2013,Crippa2023}. Although imaging is central to tumour assessment, definitive histopathological and many molecular analyses rely on cells or tissue obtained through fine-needle aspiration, core-needle biopsy, surgical biopsy, or tumour resection \cite{Tsimberidou2020}. However, surgical resection is not always feasible or clinically indicated, while fine-needle aspiration and core-needle biopsy sample only limited regions of a tumour. Consequently, these procedures are susceptible to sampling error and may not capture the full spatial extent of tumour heterogeneity  \cite{Makowska2016}.

Medical imaging provides a non-invasive, in vivo alternative for evaluating tumour status and assessing ITH. Among current imaging modalities, multiparametric magnetic resonance imaging (mp-MRI) provides complementary contrasts sensitive to regional differences in cellularity, perfusion, and microstructure. Jain et al. \cite{Jain2020} reported that mp-MRI appearance can serve as non-invasive imaging surrogates of ITH and correlate with genomic profiles. Multiple studies have demonstrated that tumours across various cancer types can be partitioned into distinct sub-regions on mp-MRI using computational approaches based on imaging-derived radiomic features and spatial molecular information, or through manual delineation by clinicians \cite{Parry2018,Stoyanova2016,Hu2023}. These sub-regions have been shown to correlate with underlying molecular characteristics. However, manual delineation is subjective, labour-intensive, and difficult to scale. Computational approaches such as unsupervised clustering offer an alternative by grouping tumour voxels with similar imaging characteristics into distinct within-tumour sub-regions. However, the resulting cluster labels are inherently arbitrary and may not correspond across patients, limiting their utility for learning consistent cross-patient representations of tumour heterogeneity.

Recent advances in deep learning offer a promising approach for tumour characterisation by learning imaging features associated with tumour morphology, molecular status, or clinical outcomes. Deep learning methods have achieved strong performance across a range of MRI clinical decision-support tasks, including brain tumour segmentation and prostate cancer detection and localisation \cite{Dorfner2025,He2023Transformers}. Zong et al., for example, developed a deep learning model for detecting and localising malignant prostate cancer using mp-MRI \cite{Zong2022}. Pasquini et al. employed a CNN model to predict IDH mutation status in patients with GBM \cite{Pasquini2021}. Cheng et al. developed a fully automated multi-task network that jointly performed glioma segmentation and IDH-genotype prediction from multimodal MRI \cite{Cheng2022}. Although this framework combined spatial delineation with molecular prediction, it learned supervised, predefined glioma regions rather than discovering and aligning imaging-derived heterogeneity phenotypes. Despite these advances, three challenges remain. First, many deep learning classification frameworks operate on whole-tumour regions, tumour-enclosing bounding boxes, or other coarse predefined regions, which may obscure spatially distinct imaging characteristics and therefore underrepresent ITH. Second, although tumour sub-regions may provide more informative heterogeneity-aware representations, existing sub-region based approaches often depend on manual expert delineation, making them subjective, labour-intensive, and difficult to scale. Third, although unsupervised clustering can automatically generate tumour sub-regions, when each tumour is partitioned independently before the resulting regions are re-clustered at the population level, the population-level step can group or relabel patient-specific regions but cannot alter their voxel membership or spatial boundaries established by the initial patient-level clustering. This may limit the learning of consistent cross-patient heterogeneity representations.

\begin{figure}
\centering
\includegraphics[width=0.9\textwidth]{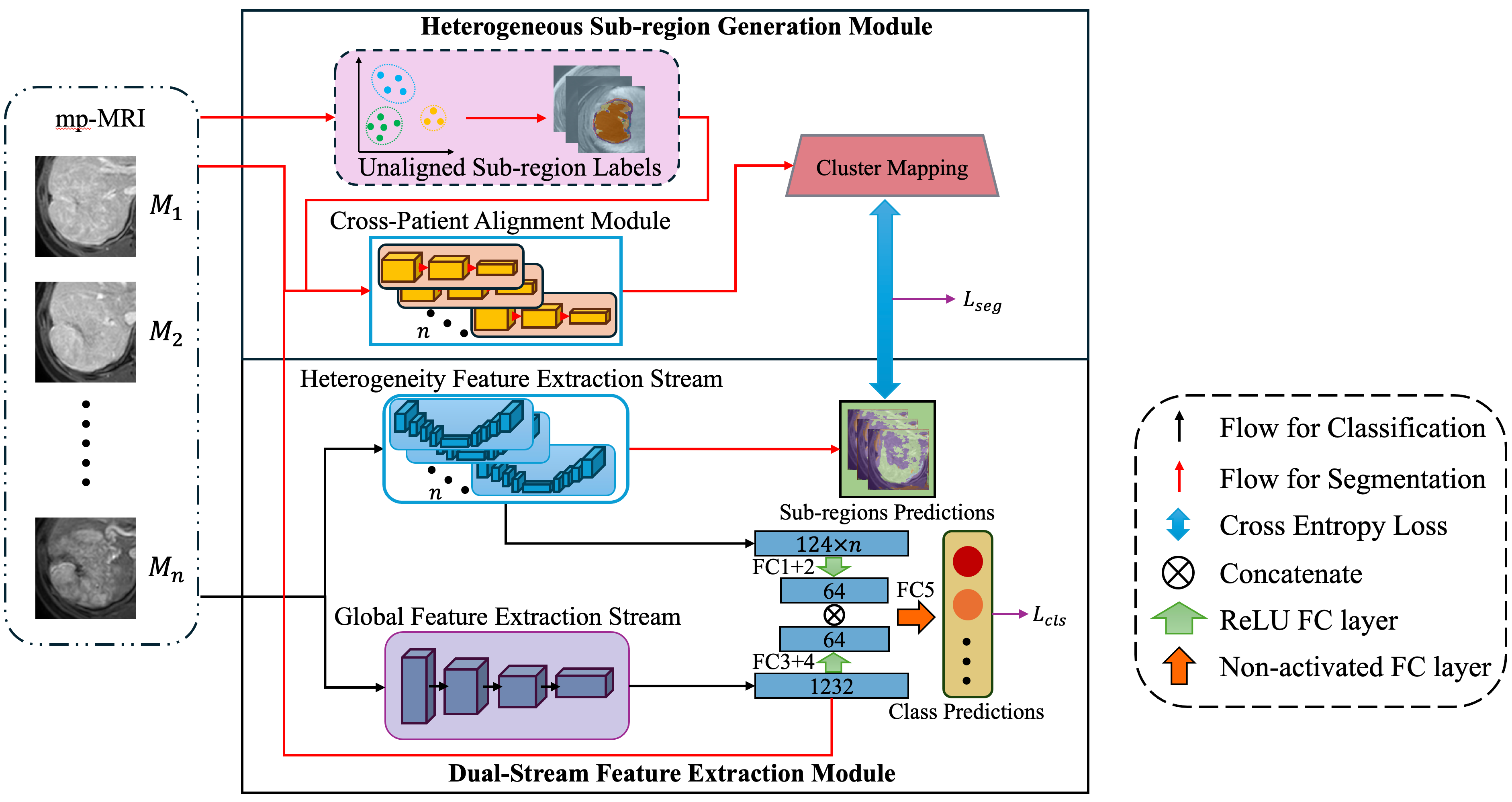}
\caption{\textbf{Overview of Heterogeneity-Aware Deep Learning Classification (HA-DLC) framework.} The framework comprises a Heterogeneous Sub-region Generation (HSG) module, Cross-Patient Sub-region Alignment (CPSA) module, and a Dual-Stream Feature Extraction (DSFE) module for heterogeneity-aware tumour classification. n denotes the number of input MRI sequences in mp-MRI. The segmentation loss ($L_{seg}$) is computed between the aligned sub-region pseudo-labels and the predicted sub-region probabilities, while the classification loss ($L_{cls}$) is computed between the ground-truth class labels and the predicted class labels.}
\label{fig1}
\end{figure}

In this study, we propose a Heterogeneity-Aware Deep Learning Classification (HA-DLC) framework that combines unsupervised sub-region generation with end-to-end heterogeneity-aware learning for tumour classification from mp-MRI. HA-DLC addresses the limitations of existing whole-tumour and manually defined sub-region approaches by automatically generating tumour sub-regions and learning cross-patient-consistent heterogeneity representations. Figure~\ref{fig1} illustrates an overview of our proposed framework. The main contributions are as follows:

\begin{itemize}
\item \textbf{Heterogeneous Sub-region Generation (HSG) Module:} We propose an unsupervised sub-region generation strategy based on clustering algorithms that operate within the tumour region of interest (ROI), producing within-tumour pseudo-labels for use in subsequent sub-region segmentation.

\item \textbf{Cross-Patient Sub-region Alignment (CPSA):} To enable unsupervised sub-regions usable for supervised heterogeneity-aware classification, we introduce a soft, learnable alignment mechanism that estimates a probability distribution mapping each patient-specific cluster to a set of shared sub-region classes. These soft correspondences convert the fixed patient-level cluster masks into aligned soft targets for supervising the sub-region segmentation networks. This reduces the effect of patient-specific cluster-label permutations and enables the classifier to learn cross-patient-comparable ITH patterns from imaging-derived sub-regions without requiring manual compartment annotations.

\item \textbf{Dual-Stream Feature Extraction (DSFE) Module:} We propose a dual-stream architecture comprising a Heterogeneity Feature Extraction (HFE) stream, which learns sequence-specific features from the generated sub-regions, and a Global Feature Extraction (GFE) stream, which captures tumour-level volumetric context. By integrating local heterogeneity information with global contextual features, DSFE produces a comprehensive representation for tumour classification. 

\end{itemize}

We evaluated HA-DLC on seven-class liver lesion diagnosis using LLD-MMRI2023 \cite{Lou2025} and MGMT promoter methylation prediction using BraTS-RC \cite{Baid2021}. HA-DLC achieved the highest mean performance among the evaluated comparison methods. Ablation experiments assessed sub-region supervision, cross-patient soft relabelling, and global tumour context. The HSG comparisons used the same outer folds without nested configuration selection and should be interpreted as post-hoc sensitivity analyses.

\section{Method}\label{sec2}

\subsection{Overview}\label{subsec:overview}

We define intra-tumoural feature heterogeneity as spatial variation in voxel-wise imaging features within the tumour ROI. For each of $N$ mp-MRI inputs (sequences or contrast phases), $V^{(n)} \in \mathbb{R}^{D \times H \times W}$, patient-level clustering produces a sub-region map $S^{(n)} \in \{0,1,\ldots,K\}^{D \times H \times W}$, where $0$ denotes background and $1,\ldots,K$ denote imaging-derived tumour clusters. CPSA maps each patient-specific cluster to a probability distribution over $M$ shared sub-region classes, producing a soft pseudo-label tensor $\tilde{\mathbf{S}}^{(n)} \in [0,1]^{(M+1)\times D\times H\times W}$. This alignment changes the cross-patient correspondence of each cluster while preserving its voxel membership and spatial boundaries. The soft-aligned pseudo-labels supervise input-specific segmentation networks, whose encoder features are fused with global tumour features for classification.

As shown in Figure~\ref{fig1}, the HFE stream within DSFE uses input-specific 3D segmentation networks to predict imaging-derived tumour sub-regions and extract local heterogeneity features from each mp-MRI input. The sub-region prediction heads are supervised by soft-aligned pseudo-label distributions generated by CPSA from the HSG clustering maps. Specifically, Lseg is the voxel-wise soft cross-entropy between the predicted sub-region distributions and the soft-aligned targets, averaged across all mp-MRI inputs. The tumour-level classification branch is supervised by Lcls,  defined as the cross-entropy loss between the predicted class distribution and the ground-truth class label. HA-DLC is optimised using the joint objective Lseg +Lcls. Together, these losses encourage the learned features to encode aligned sub-regional heterogeneity while remaining discriminative for tumour-level classification.

\subsection{Unsupervised Sub-region Label Generation}\label{subsec:hsg}

HSG generates initial patient-specific sub-region pseudo-labels within the tumour ROI using offline unsupervised clustering performed independently for each mp-MRI input. Two methods are implemented: (i) intensity-based \(k\)-means clustering \cite{Lloyd1982}, for which the number of clusters \(k\) determines the sub-region granularity; and (ii) Differentiable Feature Clustering (DFC) \cite{Kim2020}, which learns voxel features with spatial regularisation controlled by the continuity-loss coefficient \(\mu\). The resulting patient-specific label maps are subsequently aligned by CPSA.

\subsection{Cross-Patient Sub-region Alignment Module}\label{subsec:align}

The Cross-Patient Sub-region Alignment (CPSA) module softly aligns patient-specific cluster labels by estimating their correspondence to a shared set of sub-region classes. One alignment network is used for each of the $N$ mp-MRI inputs. Each network receives (i) the corresponding image volume, (ii) its raw sub-region label map generated by $k$-means or DFC, and (iii) the shared multi-scale global features from the GFE stream. Two modified DenseNet encoders \cite{Huang2017}, each comprising three dense blocks, separately encode the image volume and raw label map. The resulting image and label-map feature vectors and the globally pooled GFE feature vector are each projected to 64 dimensions and concatenated into a 192-dimensional representation. This representation is passed to $K$ cluster-specific fully connected (FC) heads, each producing $M$ assignment logits over the shared aligned classes. Thus, for patient $b$ and input $n$, the alignment network produces $\mathbf{Z}_{b,n} \in \mathbb{R}^{K \times M}$, and the complete logit tensor has shape $B \times N \times K \times M$. A softmax over the $M$ aligned classes converts each row of $\mathbf{Z}_{b,n}$ into a soft cluster-to-class probability distribution.

The soft cluster-to-class assignment probabilities are obtained as

\begin{equation}
A_{b,n,j,k}
=
\frac{\exp\!\left(Z_{b,n,j,k}\right)}
{\displaystyle\sum_{k'=1}^{M}\exp\!\left(Z_{b,n,j,k'}\right)},
\qquad
j\in\{1,\ldots,K\},\quad
k\in\{1,\ldots,M\}.
\end{equation}

$A_{b,n,j,k}$ represents the probability of mapping patient-specific cluster $j$ to aligned class $k$. The background label remains fixed and is excluded from the soft assignment. For a tumour voxel $v$ with original cluster label $j$, the soft-aligned target is defined as

\begin{equation}
\tilde{S}_{b,n,k,v}
=
A_{b,n,j,k},
\qquad
j=S_{b,n,v}\in\{1,\ldots,K\},\quad
k\in\{1,\ldots,M\}.
\end{equation}

Thus, all voxels belonging to the same initial cluster receive the same aligned-class probability distribution, preserving their voxel membership and spatial boundaries. For visualisation only, a discrete aligned sub-region map can be obtained as

\begin{equation}
\bar{S}_{b,n,v}
=
\underset{k\in\{0,\ldots,M\}}{\arg\max}\,
\tilde{S}_{b,n,k,v}.
\end{equation}

\begin{figure}
\centering
\includegraphics[width=0.9\textwidth]{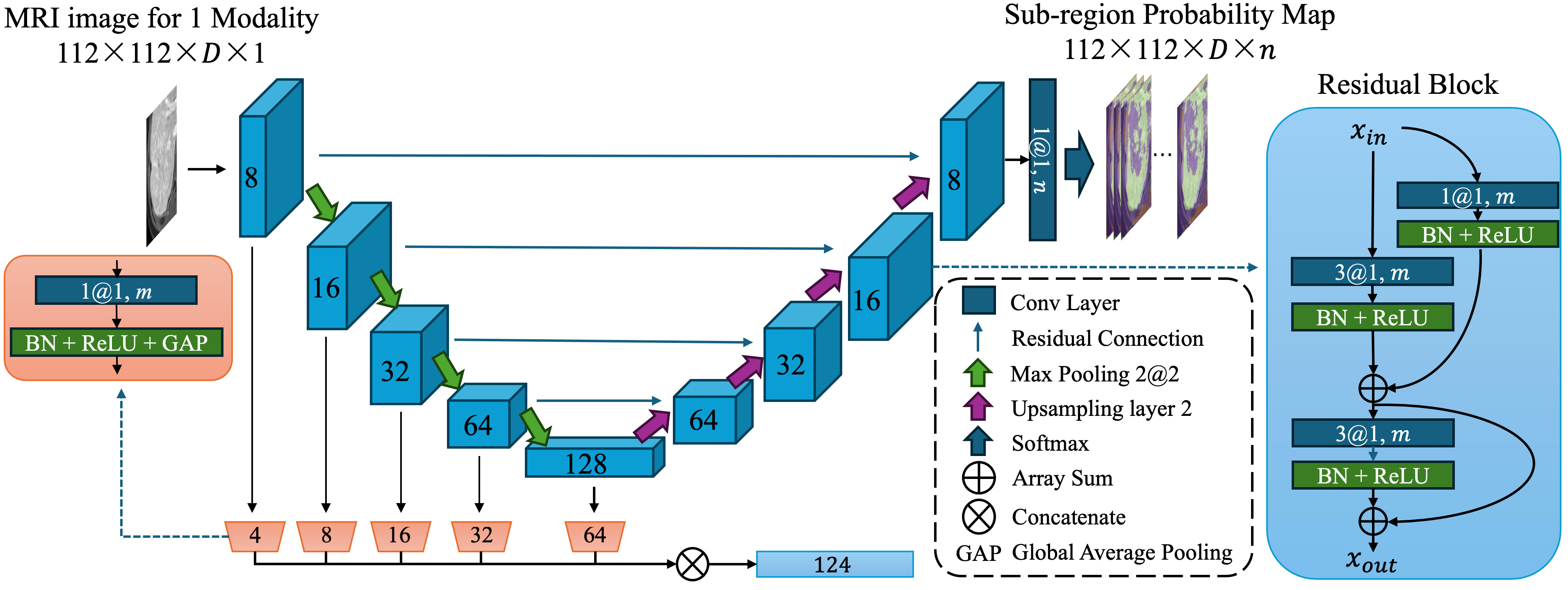}
\caption{\textbf{Heterogeneity feature extraction stream architecture.} The stream follows a 3D residual U-Net structure with encoder and decoder branches constructed from residual blocks. Local heterogeneity features are extracted from the outputs of the encoder residual blocks. The residual block denoted by $m$ contains convolutional layers with $m$ output channels. The notation $a@b,m$ denotes a kernel size of $a\times a\times a$, a stride of $b\times b\times b$, and $m$ output channels.}
\label{fig2}
\end{figure}

\subsection{Dual-Stream Feature Extraction Module}\label{subsec:dsfe}

The Dual-Stream Feature Extraction (DSFE) module extracts complementary local and global representations for heterogeneity-aware tumour classification. It comprises two streams: (i) a Heterogeneity Feature Extraction (HFE) stream, which uses an input-specific 3D segmentation network for each MRI sequence or contrast phase to predict tumour sub-regions and learn local features under soft-aligned segmentation supervision; and (ii) a Global Feature Extraction (GFE) stream, which uses a global feature backbone to capture tumour-level volumetric context from the complete mp-MRI input. The resulting local and global representations are fused to perform tumour-level classification.

\subsubsection{Heterogeneity Feature Extraction Stream}\label{subsubsec:local_hfe}

The HFE stream consists of input-specific segmentation networks, each based on a customised 3D residual U-Net architecture \cite{Diakogiannis2020}. Each network processes one mp-MRI input (sequence or contrast phase), predicts imaging-derived tumour sub-regions, and extracts local heterogeneity features from the encoder. The sub-region prediction heads are supervised using the soft-aligned pseudo-labels generated by CPSA.

As shown in Figure~\ref{fig2}, each segmentation network follows an encoder--decoder structure with skip connections. The encoder contains five cascaded residual blocks that progressively extract multi-scale local features from the input volume. Between encoder blocks, max-pooling is used to reduce spatial resolution and increase the receptive field. The decoder mirrors the encoder structure by progressively upsampling the feature maps and combining them with corresponding encoder features through skip connections. These skip connections help preserve voxel-level spatial detail for sub-region prediction.

Each residual block contains a projection branch and a main convolutional branch. The projection branch adjusts the number of feature channels so that it can be added to the output of the main branch. The main branch applies two successive convolutional operations, each followed by batch normalisation and ReLU activation. The output of the first convolutional operation is added to the projected input, and the resulting intermediate feature is added again after the second convolutional operation.

The final sub-region logits are produced by a $1\times1\times1$ convolution and converted into probabilities using softmax normalisation, where the output channels correspond to the background and shared aligned sub-region classes. In addition to segmentation, local heterogeneity features are extracted from the encoder at multiple scales. Specifically, the output of each encoder residual block is compressed using a $1\times1\times1$ convolution, followed by batch normalisation, ReLU activation, and global average pooling. The resulting multi-scale features are concatenated to form an input-specific local heterogeneity representation.

\subsubsection{Global Feature Extraction Stream}\label{subsubsec:gfe}

The GFE stream captures whole-tumour contextual information complementary to the local sub-region-aware features. We use UniFormer \cite{Li2022} as the global backbone because its hybrid convolution--attention architecture captures both local image patterns and global volumetric context. In our implementation, a UniFormer-Small-Plus configuration is used to extract the global feature representation from the mp-MRI input.

Global features are passed through fully connected layers to form a compact embedding. This embedding is concatenated with the sub-region-aware local features from all sequence-specific segmentation networks. A final classification layer uses the combined feature vector to generate tumour-level predictions. The DSFE module therefore integrates sub-region-aware local features with global tumour context.

\subsection{Loss}\label{subsec:loss}

HA-DLC is optimised using equally weighted classification and segmentation losses. HSG first generates patient- and input-specific cluster maps through offline clustering, and CPSA converts them into voxel-wise soft-aligned targets for supervising the input-specific segmentation networks. The total objective is

\begin{equation}
\mathcal{L}
=
\mathcal{L}_{\mathrm{cls}}
+
\mathcal{L}_{\mathrm{seg}}.
\label{eq:total_loss}
\end{equation}

The classification and segmentation losses are defined as

\begin{equation}
\mathcal{L}_{\mathrm{cls}}
=
\operatorname{CE}\!\left(
\mathbf{y},
\hat{\mathbf{y}}
\right),
\label{eq:classification_loss}
\end{equation}

\begin{equation}
\mathcal{L}_{\mathrm{seg}}
=
\frac{1}{N}
\sum_{n=1}^{N}
\operatorname{CE}\!\left(
\tilde{\mathbf{S}}^{(n)},
\hat{\mathbf{S}}^{(n)}
\right),
\label{eq:segmentation_loss}
\end{equation}

where $\operatorname{CE}$ denotes mean cross-entropy, $\mathbf{y}$ and $\hat{\mathbf{y}}$ are the ground-truth and predicted tumour-class distributions, and $\tilde{\mathbf{S}}^{(n)}$ and $\hat{\mathbf{S}}^{(n)}$ are the soft-aligned target and predicted sub-region distribution for mp-MRI input $n$. The segmentation loss is averaged over the batch and all voxels, including background. Because the soft targets remain differentiable functions of the CPSA assignments, $\mathcal{L}_{\mathrm{seg}}$ jointly trains CPSA and the segmentation networks and also propagates into GFE. Meanwhile, $\mathcal{L}_{\mathrm{cls}}$ supervises the local and global feature streams and the classification head. Thus, following offline clustering, all trainable components are jointly optimised.

\section{Experiment}\label{sec3}

\subsection{Dataset}\label{subsec:dataset}

The LLD-MMRI2023 dataset (LLD-M) comprises 498 patients with eight multi-phase MRI sequences: T2W, DWI, in-phase, out-of-phase, pre-contrast, arterial, portal venous, and delayed phases \cite{Lou2025}. The task is to classify seven lesion types, including four benign lesions (haemangioma, abscess, cyst, and focal nodular hyperplasia) and three malignant lesions (cholangiocarcinoma, metastases, and hepatocellular carcinoma). Whole-tumour masks provided with the dataset were generated using MedSAM within a human-in-the-loop annotation pipeline \cite{Ma2024}.

The BraTS2021 Radiogenomic Classification dataset (BraTS-RC) comprises 585 cases, with four MRI sequences: T1-weighted post-contrast (T1Gd), T1-weighted pre-contrast (T1w), T2-weighted (T2), and fluid-attenuated inversion recovery (FLAIR). To obtain tumour segmentation masks, patient identifiers were matched with the RSNA-ASNR-MICCAI Brain Tumor Segmentation Challenge 2021 dataset \cite{Baid2021}. These masks include three major tumour sub-region labels (enhancing tumour, necrotic/non-enhancing tumour core, and peritumoural oedema), whose union defines the whole tumour, and served as ground-truth labels for the expert-defined sub-region ablation. Cases without matching segmentation masks were excluded, resulting in 577 cases for analysis.

\subsection{Comparison methods}\label{subsec:comparison}

We compared HA-DLC against several state-of-the-art deep learning and radiomics-based classification methods. For the BraTS-RC dataset, comparison methods included the first-place solution from the BraTS-RC challenge \cite{Baba2021}, the Vision Graph Neural Network (ViG) framework for MGMT prediction \cite{Hu2024}, and UniFormer \cite{Li2022}. In addition, three widely used classification architectures, ResNet-50 \cite{He2016}, DenseNet-121 \cite{Huang2017}, and EfficientNet-B7 \cite{Tan2019}, were evaluated as strong 3D baselines. The first-place BraTS-RC solution employed a 3D ResNet-10 backbone and constructed a 41-slice FLAIR volume centred on the slice containing the largest brain cross-sectional area. For LLD-M, the slice index was determined from the T2W tumour mask. The ViG framework originally used fused FLAIR, T1w, and T2 pseudo-RGB images as input to a PyramidViG-S architecture. In our implementation, its input layer was adapted to incorporate all available MRI sequences: four for BraTS-RC and eight for LLD-M. For UniFormer, the UniFormer-Small-Plus architecture was implemented using the same input images as HA-DLC. As a radiomics baseline, 107 handcrafted features were extracted per ROI and MRI sequence using PyRadiomics (v3.0.1) \cite{Griethuysen2017}. These included shape descriptors, first-order intensity statistics, and texture features derived from the gray-level co-occurrence matrix (GLCM), gray-level size zone matrix (GLSZM), gray-level run length matrix (GLRLM), neighbouring gray tone difference matrix (NGTDM), and gray-level dependence matrix (GLDM). Features were extracted from either the whole-tumour region or each generated tumour sub-region across MRI sequences and subsequently used to train an XGBoost classifier \cite{Chen2016}.

\subsection{Experimental settings}\label{subsec:settings}

The provided whole-tumour masks were used to crop each tumour ROI prior to sub-region generation. HSG clustering was performed independently on each MRI sequence. For k-means clustering, the number of clusters was varied from 3 to 9. For Deep Feature Clustering (DFC), continuity-loss coefficients of 2, 0.5, and 0.001 were evaluated to control spatial continuity and the resulting sub-region granularity. A fixed random seed was used, and the best-performing configuration for each dataset was identified from the mean performance across the five folds.

We conducted two sets of ablation experiments under the same 5-fold cross-validation protocol. The first set evaluated the effect of heterogeneity supervision by comparing different segmentation targets:

\begin{enumerate}
    \item \textbf{No segmentation loss}: the segmentation branch and its local encoder features were retained, but the segmentation loss $L_{seg}$ was removed from the training objective;
    
    \item \textbf{Whole-tumour segmentation}: the whole-tumour mask was used as the segmentation target without sub-region partitioning; and
    
    \item \textbf{Ground-truth sub-region segmentation (BraTS-RC only)}: expert-defined anatomical tumour sub-regions (enhancing tumour, necrotic/non-enhancing tumour core, and peritumoural oedema) were used as the segmentation target.
\end{enumerate}

These experiments investigated whether heterogeneity-aware supervision provides additional predictive value beyond no segmentation supervision and conventional whole-tumour representations. The second set assessed the contribution of individual HA-DLC components by removing one component at a time while keeping all other settings fixed:

\begin{enumerate}
    \item \textbf{No alignment}: the CPSA module was disabled and the raw clustering labels were used directly as segmentation targets, evaluating the contribution of cross-patient sub-region alignment; and
    
    \item \textbf{Local HFE only}: only the HFE stream was used for classification, with the GFE stream removed, evaluating the contribution of global tumour context.
\end{enumerate}

These experiments were performed on both datasets to evaluate the individual contributions of CPSA and the global GFE stream within DSFE to heterogeneity-aware tumour classification.

\subsection{Evaluation metrics}\label{subsec:metrics}

For the LLD-MMRI2023 (LLD-M) dataset, the primary evaluation metric was the average of the macro-averaged F1-score and Cohen's kappa, following the challenge protocol \cite{Lou2025}. This metric was chosen to account for class imbalance by jointly assessing class-wise classification performance and agreement beyond chance. For the BraTS-RC dataset, the primary evaluation metric was the area under the receiver operating characteristic curve (AUC) \cite{Baid2021}. Accuracy was also reported as a supplementary metric to facilitate comparison across methods.

\subsection{Implementation details}\label{subsec:impl}

HA-DLC and all deep learning comparison methods were implemented in PyTorch \cite{Paszke2017}. The AdamW optimiser was used for training, with a batch size of 4 for the LLD-M dataset and 2 for the BraTS-RC dataset. To account for differences in tumour extent along the through-plane direction, all tumour ROIs were resized to $128 \times 128 \times 18$ for LLD-M and $128 \times 128 \times 36$ for BraTS-RC. Data augmentation was applied during training, including random cropping to $112 \times 112 \times 16$ and $112 \times 112 \times 32$ for LLD-M and BraTS-RC, respectively, random flipping along the three spatial axes, and random in-plane rotations of up to $10^\circ$. The initial learning rate was set to $1 \times 10^{-4}$ and was updated using the cosine learning-rate scheduler implemented in PyTorch Image Models (timm) \cite{Wightman2019}, with five warm-up epochs and a minimum learning rate of $1 \times 10^{-5}$. All deep learning experiments were conducted on a single NVIDIA RTX 4090 GPU with 24 GB of memory. Models were trained and evaluated using 5-fold cross-validation, and the reported performance corresponds to the average across the five folds.

\section{Results}\label{sec4}

\subsection{Classification Performance}\label{subsec:clsperf}

\begin{table}[htbp]
\caption{Classification performance comparison with the evaluated methods on the LLD-MMRI2023 (LLD-M) and BraTS-RC datasets. Best results are in bold and second-best are underlined.}\label{tab1}
\centering
\begin{tabular}{@{}lcccc@{}}
\toprule & \multicolumn{2}{c}{LLD-MMRI2023 (LLD-M)} & \multicolumn{2}{c}{BraTS-RC} \\
\cmidrule(lr){2-3}\cmidrule(lr){4-5}
Method                     & Accuracy          & F1--Kappa average & Accuracy          & AUC \\
\midrule
Kaggle Leaderboard         & $0.677 \pm 0.036$ & $0.629 \pm 0.048$ & $\underline{0.624} \pm 0.030$ & $0.640 \pm 0.036$ \\
ViG-S                      & $0.719 \pm 0.020$ & $0.681 \pm 0.021$ & $0.603 \pm 0.023$ & $0.647 \pm 0.023$ \\
ResNet-50                  & $0.761 \pm 0.021$ & $0.725 \pm 0.020$ & $0.622 \pm 0.031$ & $0.666 \pm 0.012$ \\
EfficientNet-B7            & $0.669 \pm 0.054$ & $0.610 \pm 0.076$ & $0.595 \pm 0.063$ & $0.657 \pm 0.021$ \\
DenseNet-121               & $0.735 \pm 0.057$ & $0.700 \pm 0.061$ & $0.603 \pm 0.023$ & $\underline{0.667} \pm 0.028$ \\
UniFormer-Small-Plus       & $\underline{0.809} \pm 0.020$ & $\underline{0.778} \pm 0.029$ & $0.592 \pm 0.019$ & $0.606 \pm 0.015$ \\
\textbf{HA-DLC (Proposed)} & $\mathbf{0.841} \pm 0.033$ & $\mathbf{0.819} \pm 0.033$ & $\mathbf{0.660} \pm 0.024$ & $\mathbf{0.686} \pm 0.036$ \\
\botrule
\end{tabular}
\end{table}

Table~\ref{tab1} presents the classification results comparing HA-DLC with state-of-the-art methods. For the LLD-M dataset, HA-DLC achieved an accuracy of 0.841, which was 0.032--0.172 higher than those of the comparison methods. In terms of the F1--kappa average, our model reached 0.819, exceeding the comparison methods by 0.041--0.209. For the BraTS-RC dataset, HA-DLC obtained an accuracy of 0.660 and an AUC of 0.686, outperforming the comparison methods by differences of 0.036--0.068 and 0.019--0.080, respectively.

Table~\ref{tab2} reports the sensitivity of HA-DLC to the HSG clustering configuration. For LLD-M, the F1--kappa average ranged from 0.791 to 0.819 across the tested configurations. K-means with $k=5$ achieved the highest value of 0.819, although several alternatives yielded numerically similar values. On BraTS-RC, the AUC ranged from 0.663 to 0.686, with DFC at $\mu=0.5$ achieving the highest value.

\begin{table}[htbp]
\caption{Sensitivity of HA-DLC to HSG clustering configuration on LLD-M and BraTS-RC. All configurations were evaluated using the same outer five-fold splits without nested inner selection. Values are cross-validation validation means and population standard deviations across folds. Best result per dataset is in bold.}\label{tab2}
\centering
\begin{tabular}{@{}llcc@{}}
\toprule
Clustering & Parameter & LLD-M (F1--Kappa average) & BraTS-RC (AUC) \\
\midrule
\multirow{7}{*}{k-means}
 & $k=3$       & $0.814 \pm 0.012$ & $0.678 \pm 0.048$ \\
 & $k=4$       & $0.805 \pm 0.042$ & $0.667 \pm 0.021$ \\
 & $k=5$       & $\mathbf{0.819} \pm 0.033$ & $0.669 \pm 0.030$ \\
 & $k=6$       & $0.799 \pm 0.036$ & $0.667 \pm 0.029$ \\
 & $k=7$       & $0.809 \pm 0.049$ & $0.670 \pm 0.024$ \\
 & $k=8$       & $0.791 \pm 0.027$ & $0.668 \pm 0.038$ \\
 & $k=9$       & $0.793 \pm 0.045$ & $0.669 \pm 0.032$ \\
\midrule
\multirow{3}{*}{DFC}
 & $\mu=0.001$ & $0.816 \pm 0.028$ & $0.663 \pm 0.030$ \\
 & $\mu=0.5$   & $0.800 \pm 0.031$ & $\mathbf{0.686} \pm 0.036$ \\
 & $\mu=2$     & $0.814 \pm 0.036$ & $0.674 \pm 0.037$ \\
\botrule
\end{tabular}
\end{table}

Table~\ref{tab3} summarises the impact of different sub-region definitions on radiomics-based classification performance for the LLD-M and BraTS-RC datasets. Three region definitions were evaluated: (i) whole-tumour masks, (ii) discrete sub-region maps predicted by the trained HA-DLC model, and (iii) expert-defined anatomical tumour sub-regions (BraTS-RC only). On the LLD-M dataset, radiomics features extracted from HA-DLC-predicted sub-regions achieved an F1--kappa average of $0.616 \pm 0.071$, numerically outperforming the whole-tumour approach ($0.565 \pm 0.036$). Similarly, for MGMT promoter methylation prediction, radiomics features derived from HA-DLC-predicted sub-regions achieved an AUC of $0.576 \pm 0.049$, numerically exceeding both the whole-tumour baseline ($0.539 \pm 0.038$) and the expert-defined anatomical tumour sub-regions ($0.558 \pm 0.037$).

\begin{table}[htbp]
\caption{XGBoost classification results using radiomics features from different tumour sub-regions.}\label{tab3}
\centering
\begin{tabular}{@{}lcc@{}}
\toprule
Sub-region used & LLD-M (F1--Kappa average) & BraTS-RC (AUC) \\
\midrule
Whole tumour    & $0.565 \pm 0.036$            & $0.539 \pm 0.038$ \\
Generated       & $\mathbf{0.616} \pm 0.071$   & $\mathbf{0.576} \pm 0.049$ \\
Expert-defined  & NA                           & $0.558 \pm 0.037$ \\
\botrule
\end{tabular}
\end{table}

\subsection{Ablation studies}\label{subsec:ablation}

Table~\ref{tab4} presents the ablation study results on the LLD-M and BraTS-RC datasets. The first set of experiments evaluated the effect of different segmentation targets and the removal of segmentation supervision on classification performance. Removing the segmentation loss reduced the F1--kappa average to 0.768 on LLD-M and the AUC to 0.623 on BraTS-RC. Using the whole-tumour mask as the segmentation target improved the corresponding scores to 0.802 and 0.650, respectively. Replacing whole-tumour supervision with soft, aligned imaging-derived sub-region supervision further improved performance to 0.819 on LLD-M and 0.686 on BraTS-RC. On BraTS-RC, using expert-defined anatomical tumour sub-regions as the segmentation target achieved an AUC of 0.681, which was slightly lower than that obtained using the proposed imaging-derived sub-regions.

The second set of experiments evaluated the contribution of individual HA-DLC components. Disabling the CPSA module and directly using the raw patient-specific clustering labels as segmentation targets reduced the F1--kappa average to 0.784 on LLD-M and the AUC to 0.618 on BraTS-RC, supporting the contribution of cross-patient sub-region alignment. Restricting the classifier to the HFE stream alone resulted in an F1--kappa average of 0.769 on LLD-M and an AUC of 0.669 on BraTS-RC, indicating that incorporating global tumour context through the GFE stream improved the mean classification performance.

\begin{table}[htbp]
\caption{Ablation study on LLD-M and BraTS-RC (5-fold cross-validation). The upper block varies the segmentation target; the lower block removes one component of HA-DLC at a time. $\Delta$ is relative to the full HA-DLC. Best result per dataset in bold.}
\label{tab4}
\centering
\begin{tabular}{@{}lcccc@{}}
\toprule
 & \multicolumn{2}{c}{LLD-M (F1--Kappa average)} & \multicolumn{2}{c}{BraTS-RC (AUC)} \\
\cmidrule(lr){2-3}\cmidrule(lr){4-5}
Variant & Score & $\Delta$ & Score & $\Delta$ \\
\midrule
HA-DLC (full) & $\mathbf{0.819} \pm 0.033$ & --- & $\mathbf{0.686} \pm 0.036$ & --- \\
\midrule
\multicolumn{5}{@{}l}{\textit{Segmentation target}}\\
\quad w/o segmentation loss      & $0.768 \pm 0.029$ & $-0.051$ & $0.623 \pm 0.054$ & $-0.063$ \\
\quad Whole-tumour segmentation  & $0.802 \pm 0.031$ & $-0.017$ & $0.650 \pm 0.021$ & $-0.036$ \\
\quad Expert-defined sub-regions  & NA                & ---      & $0.681 \pm 0.022$ & $-0.005$ \\
\midrule
\multicolumn{5}{@{}l}{\textit{Component removed}}\\
\quad w/o alignment         & $0.784 \pm 0.047$ & $-0.035$ & $0.618 \pm 0.016$ & $-0.068$ \\
\quad Local HFE only             & $0.769 \pm 0.045$ & $-0.050$ & $0.669 \pm 0.036$ & $-0.017$ \\
\botrule
\end{tabular}
\end{table}

\subsection{Qualitative and Quantitative Analysis of Predicted Sub-regions}\label{subsec:segeval}

\begin{figure}[htbp]
\centering
\includegraphics[width=0.75\textwidth]{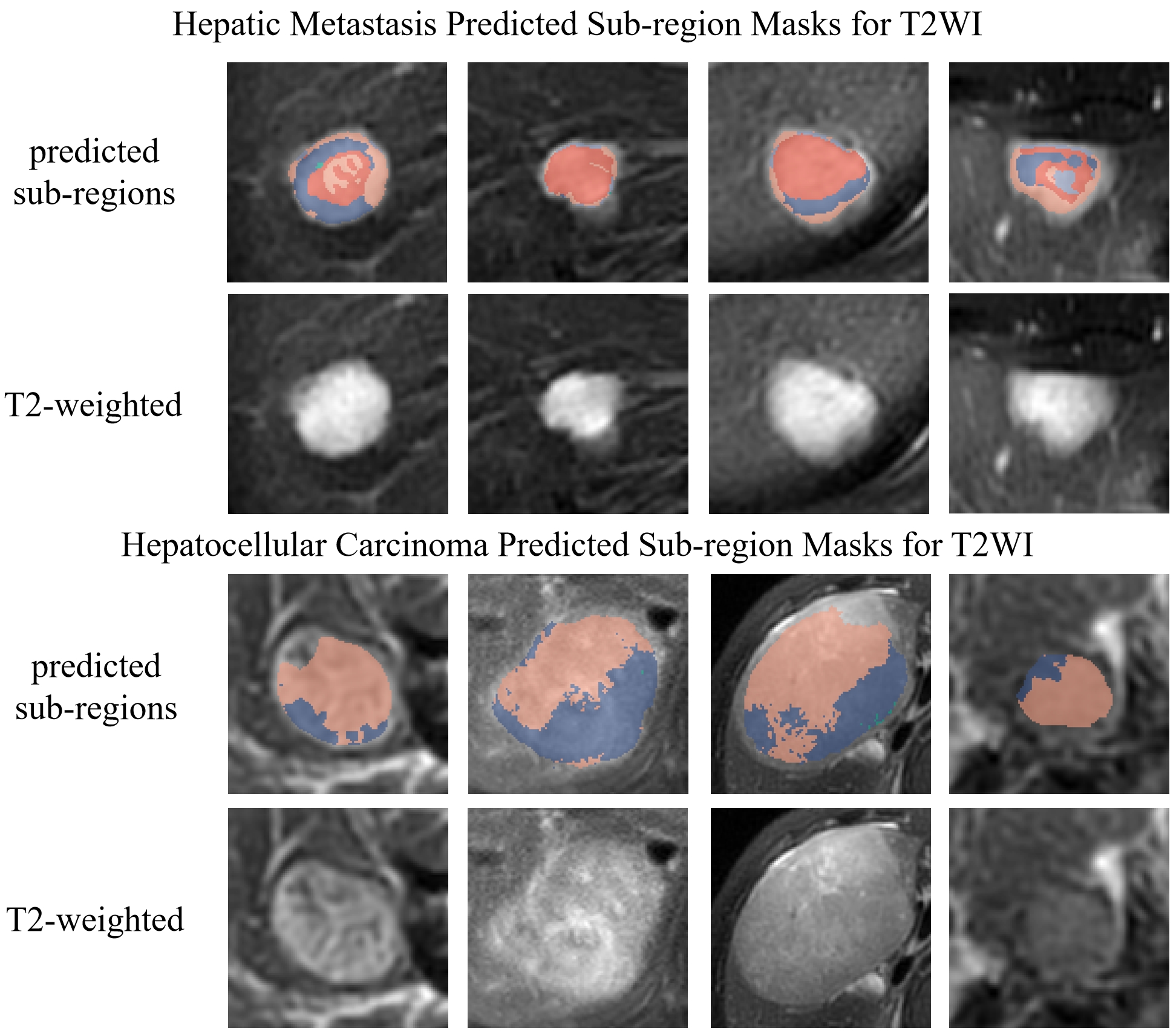}
\caption{\textbf{Sub-region segmentation of liver tumour on T2-weighted MRI.} The top two rows show four representative hepatic metastases, with predicted sub-region maps in the first row and the corresponding T2-weighted images in the second row. The bottom two rows show four representative hepatocellular carcinomas (HCC), with predicted sub-region maps in the third row and the corresponding T2-weighted images in the fourth row. The predicted sub-regions are overlaid in semi-transparent colours, illustrating lesion-specific intra-tumoural heterogeneity patterns identified by HA-DLC.}
\label{fig3}
\end{figure}

Figure~\ref{fig3} illustrates the sub-region maps predicted by HA-DLC on T2-weighted images for four representative cases of hepatic metastases and four cases of hepatocellular carcinoma (HCC). The colour-coded maps suggest that the aligned sub-regions capture recurring heterogeneity patterns across patients. In metastatic lesions, HA-DLC identified distinct sub-region structures that often corresponded to intensity zones visible on the T2-weighted images, indicating heterogeneous imaging characteristics within the lesions. In contrast, HCC lesions exhibited a different sub-region organisation, with one sub-region occupying much of the tumour interior and another concentrated near the lesion boundary. These observations suggest that the proposed framework may identify lesion-type-associated spatial imaging patterns. Importantly, the generated sub-regions were derived entirely from imaging data without histopathological supervision; therefore, the biological significance of individual clusters requires further validation.

\begin{figure}[htbp]
\centering
\includegraphics[width=\columnwidth]{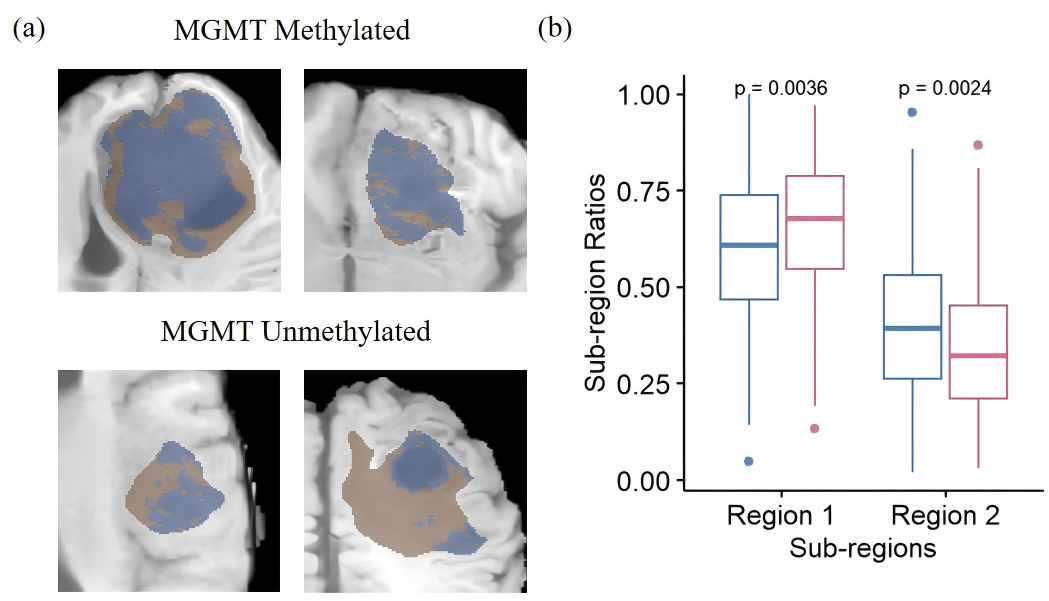}
\caption{\textbf{Sub-region segmentation and quantitative comparison by MGMT promoter methylation status.} (a) Representative T1-weighted MRI slices of glioblastoma lesions from MGMT-methylated tumours (top row) and MGMT-unmethylated tumours (bottom row). Sub-region 1 is shown in blue and Sub-region 2 in brown. (b) Boxplots of normalised sub-region volume ratios (sub-region volume/total tumour volume), stratified by methylation status (blue boxes: unmethylated; red boxes: methylated).}
\label{fig4}
\end{figure}

The learned sub-regions were also associated with molecular status in glioblastoma. As shown in Figure~\ref{fig4}, the heterogeneity-aware framework delineated two dominant complementary regions on T1-weighted images. The normalised volume fraction of Sub-region~1 was significantly higher in MGMT-methylated tumours than in unmethylated tumours (Welch's two-sample \textit{t}-test, $n=578$, $p=0.0036$), whereas Sub-region~2 was significantly lower in the methylated cohort ($p=0.0024$). These quantitative differences were consistent with the qualitative examples shown in Figure~\ref{fig4}, which illustrate different spatial distributions of the two sub-regions according to MGMT methylation status. Together, these findings suggest that the imaging-derived sub-regions identified by HA-DLC may capture tumour characteristics associated with underlying molecular phenotypes.

\section{Discussion}\label{sec5}

Our main findings are as follows: (i) HA-DLC consistently outperformed state-of-the-art radiomics and deep learning methods on both the LLD-M and BraTS-RC datasets, supporting the effectiveness of explicitly modelling tumour heterogeneity for both lesion classification and molecular-status prediction; (ii) performance was relatively robust to the choice of HSG clustering strategy, although the optimal clustering configuration differed between datasets, suggesting that the framework can accommodate different patterns of tumour heterogeneity; (iii) imaging-derived sub-region segmentation yielded superior classification performance compared with whole-tumour segmentation and, on BraTS-RC, achieved performance comparable to expert-defined tumour compartments, highlighting the value of data-driven heterogeneity representations; and (iv) the ablation studies demonstrated that the segmentation objective, CPSA module, and GFE stream within the DSFE architecture each contributed to the overall performance gains. Collectively, these findings suggest that learning cross-patient-consistent sub-regional representations of tumour heterogeneity provides complementary information beyond conventional whole-tumour analysis and can improve tumour classification from mp-MRI.

HA-DLC achieved superior performance to all evaluated comparison methods for both liver lesion classification and MGMT promoter methylation prediction (Table~\ref{tab1}). Previous radiomics studies have demonstrated the value of incorporating tumour heterogeneity through sub-region analysis to improve classification performance \cite{Lin2021}. Consistent with these observations, our results suggest that imaging-derived sub-regions capture local heterogeneity patterns that complement information contained in whole-tumour representations. This interpretation is supported by the radiomics analysis (Table~\ref{tab3}), where features extracted from HA-DLC-predicted sub-regions consistently outperformed features extracted from whole-tumour regions and, on BraTS-RC, achieved slightly higher performance than expert-defined tumour compartments. Together, these findings indicate that explicitly modelling spatially organised imaging phenotypes within a tumour may preserve heterogeneity-related information that is particularly relevant for classification.

The clustering sensitivity analysis (Table~\ref{tab2}) indicates that HA-DLC is relatively robust to the initial HSG configuration. On LLD-M, performance varied only slightly across clustering settings, suggesting that multiple cluster configurations can generate informative sub-region candidates. Although the performance differences across BraTS-RC configurations were also modest, BraTS-RC showed a numerical tendency towards coarser partitions, suggesting that overly fine clustering could fragment task-relevant imaging patterns. Because the optimal configuration differed between datasets, clustering parameters were selected separately for each task. Nevertheless, the similar best-case performance achieved by k-means and DFC suggests that the effectiveness of HA-DLC is not dependent on a specific clustering algorithm. Instead, the primary benefit appears to arise from identifying, aligning, and learning from imaging-derived heterogeneous tumour regions.

Figures~\ref{fig3} and~\ref{fig4} illustrate recurring heterogeneity patterns learned by HA-DLC from imaging-derived tumour sub-regions. Despite substantial inter-patient variability in tumour appearance, corresponding sub-regions occupied similar relative tumour locations across patients, suggesting that CPSA can establish cross-patient sub-region correspondence. In BraTS-RC, MGMT promoter methylation status was associated with significant differences in sub-region composition (Figure~\ref{fig4}), indicating that the learned sub-regions capture information beyond global tumour morphology and intensity characteristics. Similarly, the recurring sub-region patterns observed in liver lesions (Figure~\ref{fig3}) suggest that different tumour types exhibit distinct heterogeneity signatures. Together, these findings indicate that HA-DLC can identify task-relevant imaging heterogeneity patterns without requiring manual sub-region annotations. However, the biological significance of the generated sub-regions remains to be validated through histopathological or radiogenomic studies.

The ablation results (Table~\ref{tab4}) further support the contribution of each evaluated component of HA-DLC. The segmentation-target ablation demonstrated a progressive improvement in performance when moving from no segmentation supervision to whole-tumour supervision and subsequently to imaging-derived sub-region supervision. On LLD-M, the F1--kappa average increased from 0.768 without segmentation loss to 0.802 with whole-tumour supervision and further to 0.819 with sub-region supervision. Similarly, on BraTS-RC, the AUC increased from 0.623 to 0.650 and then to 0.686. Notably, imaging-derived sub-region supervision achieved performance comparable to that obtained using expert-defined anatomical BraTS tumour compartments (AUC: 0.686 versus 0.681), suggesting that effective heterogeneity-aware supervision can be obtained without manual sub-region annotation. The component-removal experiments further demonstrated the complementary roles of CPSA and the GFE stream within DSFE. Incorporating CPSA increased performance from 0.784 to 0.819 on LLD-M and from 0.618 to 0.686 on BraTS-RC, highlighting the importance of establishing consistent sub-region identities across patients. Similarly, combining the Global Feature Extraction (GFE) stream with the Heterogeneity Feature Extraction (HFE) stream improved performance from 0.769 to 0.819 on LLD-M and from 0.669 to 0.686 on BraTS-RC, demonstrating that local heterogeneity patterns and global tumour context provide complementary information for classification. Interestingly, the relative contribution of these components differed between datasets. The larger performance gain associated with GFE on LLD-M suggests that global tumour context may be particularly important for distinguishing liver lesion types, whereas the greater contribution of CPSA on BraTS-RC suggests a stronger dependence on consistent local heterogeneity patterns for MGMT promoter methylation prediction. Overall, these findings indicate that heterogeneity-aware sub-region modelling, cross-patient alignment, and multi-scale feature learning jointly contribute to improved tumour classification.

Despite the promising results, we note several limitations. First, the generated sub-regions depend on the initial clustering and alignment processes. Although the imaging-derived sub-regions achieved performance comparable to expert-defined tumour compartments on BraTS-RC, they do not necessarily correspond to established anatomical or pathological regions. This suggests that the most effective representation of tumour heterogeneity for classification may be task-dependent and does not always align with conventional tumour compartment definitions. Consequently, the biological interpretation of the learned sub-regions remains limited without histopathological or radiogenomic validation. Second, although HA-DLC demonstrated relatively stable performance across different clustering configurations (Table~\ref{tab2}), the optimal clustering parameters varied between datasets. The current framework therefore requires dataset-specific selection of clustering hyperparameters, including the number of clusters for k-means and the continuity-loss coefficient for DFC. Future work will investigate adaptive strategies for automatically determining suitable clustering configurations while reducing computational overhead. Finally, the proposed framework was evaluated on two public mp-MRI datasets representing different tumour classification tasks. Although the consistent performance improvements observed across both datasets suggest potential generalisability, the evaluation relied on internal five-fold cross-validation; therefore, validation on larger independent multi-centre cohorts, additional tumour types, and other imaging modalities is still required.

\section{Conclusion}\label{sec6}

In this study, we introduced a Heterogeneity-Aware Deep Learning Classification (HA-DLC) framework for tumour classification from mp-MRI. HA-DLC combines unsupervised imaging-derived sub-region generation, cross-patient sub-region alignment, and dual-stream feature extraction to jointly capture intra-tumoural heterogeneity and global tumour context. By generating and softly aligning sub-region pseudo-labels, the framework enables heterogeneity-aware representation learning without requiring manual sub-region annotations. Experiments across two distinct MRI datasets demonstrated that HA-DLC consistently outperformed the comparison methods, supporting the value of incorporating imaging-derived heterogeneity into deep learning-based tumour classification. These findings indicate that modelling structured intra-tumoural variation can provide complementary information for both lesion-type classification and molecular-status prediction.

\bibliography{sn-bibliography}

\end{document}